\pdfoutput=1

\documentclass[11pt]{article}

\usepackage[preprint]{acl}

\usepackage{times}
\usepackage{latexsym}

\usepackage[T1]{fontenc}

\usepackage[utf8]{inputenc}

\usepackage{microtype}

\usepackage{inconsolata}

\usepackage{url}            
\usepackage{booktabs}       
\usepackage{amsfonts}       
\usepackage{nicefrac}       
\usepackage{xcolor}         
\usepackage{graphicx}         
\usepackage{amsmath}         
\usepackage{booktabs} 
\usepackage{pifont}
\usepackage{hyperref}
\usepackage{threeparttable}
\usepackage{multirow}
\title{Refine Large Language Model  Fine-tuning via Instruction Vector}

\author{
  Gangwei Jiang\textsuperscript{1},
  Zhaoyi Li\textsuperscript{1},
  \textbf{Defu Lian\textsuperscript{1\dag}},
  \textbf{Ying Wei\textsuperscript{2}}\thanks{\hspace{1mm} Corresponding author.} \\  
  \textsuperscript{1}University of Science and Technology of China ,
  \textsuperscript{2}Nanyang Technological University  \\ 
  \texttt{gwjiang@mail.ustc.edu.cn} \\
}

\begin{document}
\maketitle

\begin{abstract}

Fine-tuning large language models (LLMs) can cause them to lose their general capabilities. However, the intrinsic mechanisms behind such forgetting remain unexplored. In this paper, we begin by examining this phenomenon by focusing on knowledge understanding and instruction following, with the latter identified as the main contributor to forgetting during fine-tuning. 
Consequently, we propose the Instruction Vector (IV) framework to capture model representations highly related to specific instruction-following capabilities, thereby making it possible to understand model-intrinsic forgetting. 
Through the analysis of IV dynamics pre and post-training, we suggest that fine-tuning mostly adds specialized reasoning patterns instead of erasing previous skills, which may appear as forgetting.
Building on this insight, we develop IV-guided training, which aims to preserve original computation graph, thereby mitigating catastrophic forgetting. Empirical tests on three benchmarks confirm the efficacy of this new approach, supporting the relationship between IVs and forgetting. Our code will be made available soon.

\end{abstract}


\section{Introduction}

Instruction fine-tuning~\citep{peng2023instruction, chung2024scaling} has emerged as an indispensable ingredient in the development of Large Language Models (LLMs)~\citep{brown2020language,radford2019language,touvron2023llama},enabling them to meet the demands of specific domains~\cite{roziere2023code, thirunavukarasu2023large} and human preferences~\cite{ouyang2022training}. However, a notable concern with this fine-tuning is "catastrophic forgetting"~\citep{mccloskey1989catastrophic, Kirkpatrick_2017}, where models may lose essential skills~\citep{dou2023loramoe,chen2023chatgpt} such as mathematical reasoning while adjusting to user instructions. This raises questions about which abilities are most susceptible to forgetting and the underlying causes of these losses in LLMs.

\begin{figure}[t]
  \centering
  \includegraphics[width=1.\linewidth]{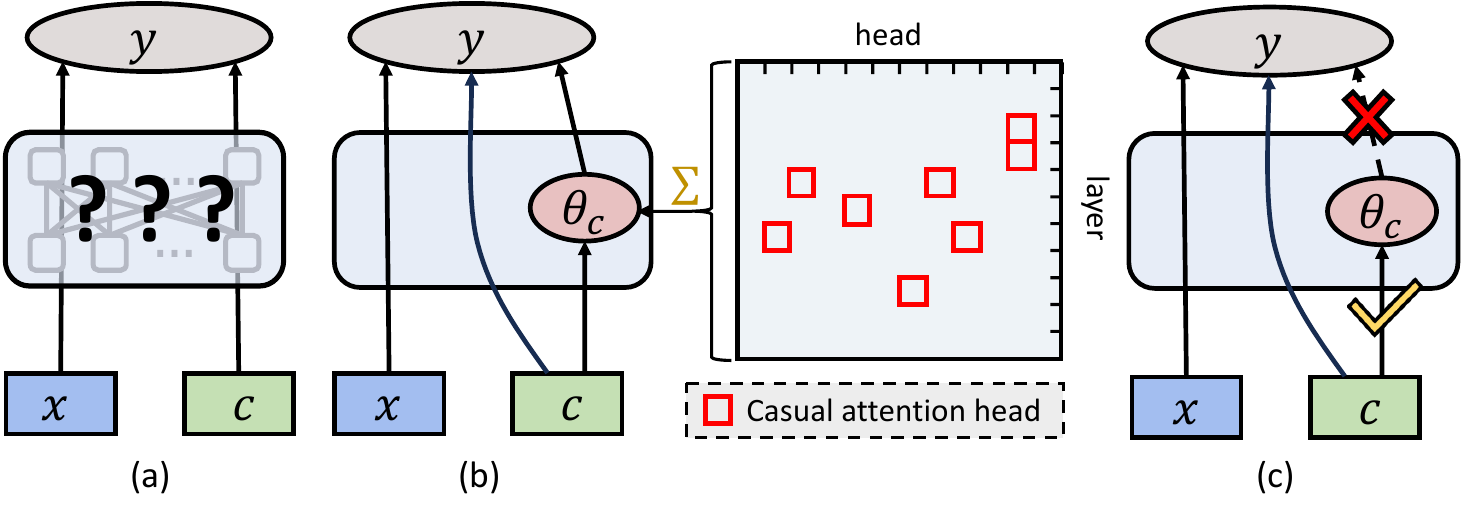}
  \vspace{-0.6em}
  \caption{Instruction vector hypothesis for LLM understanding. $\theta_c$ is extracted by aggregating representations of attention heads identified to have causal influence to the output. Forgetting is resulted from the suppression of instruction vector associated computation graph.}
  \label{fig:sec1:hypothesis}
  \vspace{-0.6em}
\end{figure}

Research on LLM forgetting~\citep{luo2024empirical, wang2023trace, wu2024llama} generally 
examines changes in abilities like reading comprehension, factual retention, mathematical skills, and code generation, underscoring the existence of catastrophic forgetting. Despite these findings, there is a notable gap in understanding the internal mechanisms responsible for these losses. To date, only a few studies, such as~\citet{kotha2024understanding} proposing the task inference hypothesis, have begun to explore how conflicts between task processors might lead to forgetting. Nevertheless, the literature still lacks comprehensive insights into the exact changes that result in forgetting, leaving open questions about whether these changes involve overwriting of old modules or if they are simply overshadowed by new, specialized patterns.

In this paper, we first present a novel perspective to investigate catastrophic forgetting in LLMs, focusing on the capabilities developed during pre-training and alignment phases. We suggest that the task proficiency in LLMs involves understanding task-specific knowledge and following instructions, assessed through \textit{Knowledge Probability} $P(y|x)$ and \textit{Instruction Probability} $P(y^c|c, x)$, respectively (as depicted in Fig.~\ref{fig:sec2:case}).
Our empirical analysis within a continual instruction tuning framework reveals distinct forgetting patterns between these two aspects, with shifts in instruction following primarily driving performance declines.



To investigate the internal changes of the model during forgetting, we introduce the Instruction Vector (IV) framework to extract representations closely associated with the task processing.
We hypothesize a straightforward yet robust computational graph for LLMs (see Fig.~\ref{fig:sec1:hypothesis} b), featuring an intermediate variable $\theta_c$ crucial for task performance. The presence or absence of $\theta_c$ directly impacts the model's capability to handle instruction $c$. This hypothesis is supported by causal intervention experiments in  Sec.~\ref{sec3.2}. By analyzing IV dynamics pre and post-training, we find minor changes in IV expression with forgetting happens. Furthermore, explicitly incorporating IV into the model's computational graph can recover the mastery of the corresponding instruction. This results indicate that fine-tuning mostly adds specialized reasoning patterns instead of erasing previous skills, which may appear as forgetting.


Building on these insights, we develop  an IV-guided training methodology to mitigate catastrophic forgetting. This method incorporates a progressive IV-intervention training mechanism, in which the IV is initially introduced through intervention and is then gradually phased out during the training process. The deliberate inclusion of IV aids in optimizing the model by ensuring adherence to the IV-related computational graph, thereby minimizing the overshadowing effect of new reasoning pathways. Additionally, we have introduced an IV-based KL-Divergence loss function to reduce the discrepancies between zero-shot and IV-intervened logits, ensuring that the model's behavior remains aligned with the original computational structure. Validated across multiple datasets, this method significantly alleviate forgetting in both general and in-context learning abilities, confirming the link between IV and forgetting.

\textbf{Main Findings and Contributions.}
\textbf{(1)} We introduce a new perspective on catastrophic forgetting by using Knowledge and Instruction Probability to evaluate how well LLMs retain task-specific knowledge and follow instructions after tuning, showing that changes in instruction adherence mainly drive performance declines.
\textbf{(2)} We are the first to interpret forgetting with the Instruction Vector framework, identifying inherent changes during fine-tuning. The findings indicate that fine-tuning generally introduces specialized reasoning patterns rather than removing existing skills.
\textbf{(3)} We develop an IV-guided training approach that focuses on preserving and realigning the model’s computational graph during fine-tuning. This significantly enhances the general and in-context learning capabilities across various datasets in continual learning.



\section{Catastrophic Forgetting in LLMs}
\label{sec2}
\begin{figure}[b]
  \centering
  \includegraphics[width=0.9\linewidth]{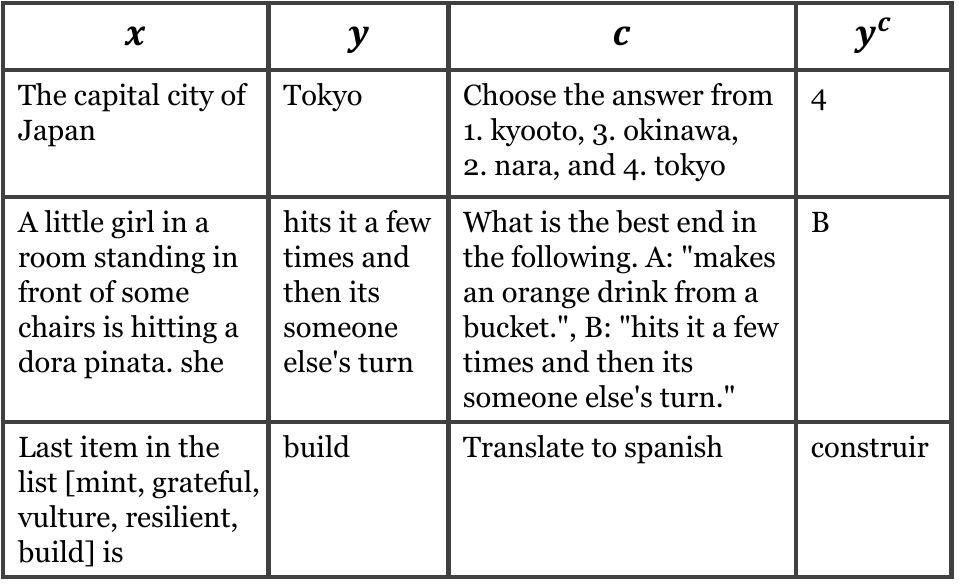}
  \vspace{-0.3em}
  \caption{Task in world knowledge form $(x,y)$ and instruction form $(x,c,y^c)$.}
  \label{fig:sec2:case}
\end{figure}

In this section, we present a new perspective to investigate catastrophic forgetting in LLMs, concentrating on the capabilities embedded within pre-training and instruction tuning stages, as opposed to focusing on pure performance shifts as noted in earlier studies~\citep{wang2023trace, zhai2023investigating}. We start with a discussion on the capabilities encoded in LLMs, proceed to develop continual instruction tuning setup to investigate forgetting, and conclude with the empirical observations.



Let $M$ denote the model pre-trained on large scale data corpus $\mathcal{D}_{PT} = \{\boldsymbol{X}_i\}$ with the language modeling task~\citep{brown2020language, radford2019language}. We assume that $M$ has built an impressive ability to capture world knowledge across various domains, i.e., $M$ assigns the maximum likelihood to $P(y|x, M)$ for certain datasets denoted by $D^{K}=\{(x_i,y_i)\} \in \mathcal{D}_{PT}$. Here, the pair $[x_i,y_i]$ may represent a segment extracted from raw text $\boldsymbol{X}_j$. For example, consider $x$ being "The capital city of Japan is" and $y$ being "Tokyo"; such a pairing frequently appears in blogs. In this paper, we refer to $P(y|x, M)$ as the \textbf{\textit{Knowledge Probability}}, which serves as a metric for evaluating the model's proficiency in comprehending world knowledge.

While processing instructional data, the model $M$ is presented with the dataset $D^{c}=\{(c, x_i, y^c_i)\}$, where each tuple consists of an instruction $c$, an input prompt $x_i$, and an expected output $y^c_i$. For instance, $c$ might be "Choose the best answer from A, B, C, and D (with options given).", $x$ could be "The capital city of Japan is", and $y^c$ would be "D", which aligns with the answer "Tokyo". The model is supposed to generate $y^c$ that accurately responds to the instruction $c$ with the context of $x$, i.e., maximize $P(y^c|c, x, M)$, which is termed as the \textbf{\textit{Instruction Probability}}. 

In this paper, when discussing catastrophic forgetting of a task, we consider alterations in both \textit{Knowledge} and \textit{Instruction Probabilities}. Typically, a test instance $x_i$ is typically presented as a tuple $(x_i, y_i, c,y^c_i)$ (examples are listed in Fig.~\ref{fig:sec2:case}), with shifts in $P(y_i^c|c, x_i, M)$ signaling variations in the model's proficiency in instruction processing and knowledge understanding and shifts in $P(y_i|x_i, M)$ solely reflect changes in the world knowledge comprehension. 
Our work go beyond simple performance metrics evaluation, offering a detailed examination of distinct capabilities amidst CF. This method reveals if performance degradation stems from an actual loss of world knowledge or a reduction in the ability to follow instructions.




\paragraph{Continual instruction tuning setup.} To explore CF in LLMs, we conduct an empirical study within the continual instruction tuning framework. In this setup, a model is sequentially trained on a series of streaming tasks, denoted as $\{D^{c_1}, D^{c_2}, ..., D^{c_T}\}$. 
Here, $D^{c_t} = \{(c_t, x_i, y^c_i)\}$ symbolizes the t-th task associated with a specific instruction $c_t$. While learning each task $D^{c_t}$, the model can only access to the corresponding data, with the goal of minimizing loss on all learned tasks. Specifically, the model is optimized with $\min_{M} \frac{1}{N} \sum_{i=1}^{N} \ell(y_i, M(c, x_i))$, where $N$ is the size of training set and $\ell$ is usually the cross-entropy loss on the entire vocabulary.
In addition to avoiding forgetting on previous learned tasks $\{D^{c_1},..., D^{c_{t-1}}\}$, the model is also evaluated on held-out evaluation sets (e.g., CommonsenseQA~\citep{talmor2018commonsenseqa},  MMLU~\citep{hendrycks2020measuring}) to measure its general ability.

We select two different continual instruction tuning benchmarks. The first is from TRACE~\citep{wang2023trace} benchmark, which consists of 6 different complex generation tasks including multi-choice QA, code generation, mathematical reasoning and summary. The second is called FUNC, adapted from the datasets in ~\citet{todd2023function}, in which tasks have clear and simple instructions. For example, task Verb-Spanish and Last-Spanish are both translation task but differ in the selection from list. For the general evaluation datasets, we utilize Hellaswag~\citep{zellers2019hellaswag}, ARC-challenge~\citep{clark2018think}, CommonsenseQA~\citep{talmor2018commonsenseqa}, and MMLU-social~\citep{hendrycks2020measuring}. The detailed dataset information and evaluation metrics are present in Appendix~\ref{app:dataset}.

We adopt LLAMA2-7B-Chat~\citep{touvron2023llama} as the base model, with its effectiveness in both understanding world knowledge and following instructions. Without specific notification, the model is fine-tuned with LORA approach~\cite{hu2021lora}, using the Adam optimizer with a learning rate set to 1e-4. Additional details regarding the implementation are provided in the Appendix~\ref{app:implement}.



\begin{figure*}[ht]
  \centering
  
  \vspace{-0.5em}
  \includegraphics[width=.99\linewidth]{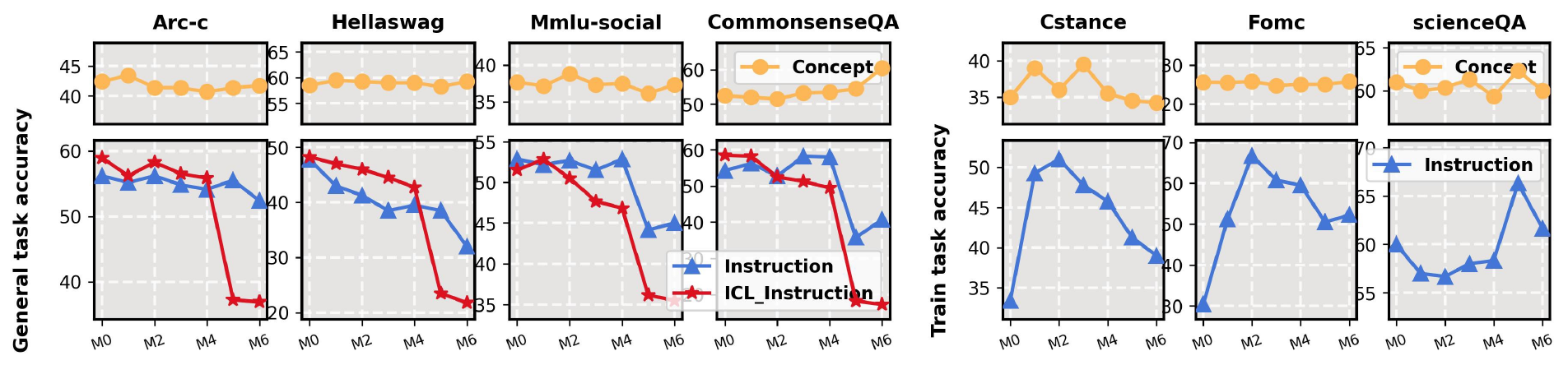}
  \caption{Accuracy curve across naive sequential instruction fine-tuning on the TRACE benchmark. X-axis delineates the stages through training, with "M0" indicating the original pre-trained model, and "Mi" signifying the model post-instruction fine-tuning for the i-th task in sequence. The tasks follow the sequence of Cstance, Fomc, Meetingbank, Py150, ScienceQA, and Numgluecm. Y-axis indicates the rank classification accuracy. Notably, the first four datasets are absent from the training set, whereas the final three datasets are part of the training distribution.}
  \label{fig:sec2:acc_curve}
\end{figure*}

\paragraph{Forgetting properties in knowledge and instruction probabilities.}

In our empirical study, we aim to investigate the factors responsible for the model performance drop. To show this, we present the accuracy curve for task in knowledge and instruction forms (cases in Fig.~\ref{fig:sec2:case}) during continual tuning in Fig.~\ref{fig:sec2:acc_curve}. Knowledge accuracy is determined by evaluating $P(y|x)$, whereas instruction accuracy is derived from $P(y^c|c, x)$. The reported accuracy follows the evaluation method in ~\citet{brown2020language, bordes2016learning} which involves choosing the label with the highest log-likelihood. The results reveal a consistent presence of the forgetting effect in LLMs across both general and newly acquired tasks throughout continual instruction tuning. More observations are as follow:

1) \textit{Instruction Following Accuracy Decline}. At the end of training sequence, the average instruction accuracy for the general evaluation set decreases by 10.24 as compared to the pre-trained model. On the other hand, knowledge accuracy sees an average increase of 1.93. This suggests loss in instruction following ability is the reason for task performance drop. 2) \textit{In-Context Learning (ICL) Ineffectiveness}: When attempting to recover performance with ICL (see the red line in Fig.~\ref{fig:sec2:acc_curve}), we observe a average decrease of 14.67 in performance compared to zero-shot results. The significant decline indicates that the bias in instruction-following ability is further magnified by ICL. 3) \textit{Severe Forgetting of Newly Learned Concepts}: Forgetting of newly acquired skills is particularly significant. The drop in results for Cstance reaches as much as 3.0 points at each stage of training, while in tasks like ARC the number is just 0.63.  

\section{Interpret Catastrophic Forgetting via Instruction Vector}

Our empirical research indicates that, during the tuning process, models tend to forget instruction-following capabilities as opposed to world knowledge understanding aptitudes. To further investigate the inherent mechanisms of such forgetting, we introduces a framework for interpretability, utilizing Instruction Vectors (IV) to decouple the distinct functionalities of the model. This approach is inspired by the ideas presented by \citet{todd2023function} and \citet{hendel2023context}, which suggest that an input-output function can be represented as a vector within LLMs. We reveal that the activation level of IV is positively correlated with the LLMs’ proficiency in relevant instruction-following skills during training. Through the analysis of IV's consistency before and after instruction tuning, this paper elucidates the fundamental mechanisms of forgetting within LLMs. 

Subsequently, we will first put forth our hypothesis and then introduce the Instruction Vectors framework. Finally, displaying the experimental results on IV, unveiling the dynamic process of forgetting.

\subsection{Instruction Vector Hypothesis}

\begin{figure*}[ht]
  \centering
  \includegraphics[width=1.\linewidth]{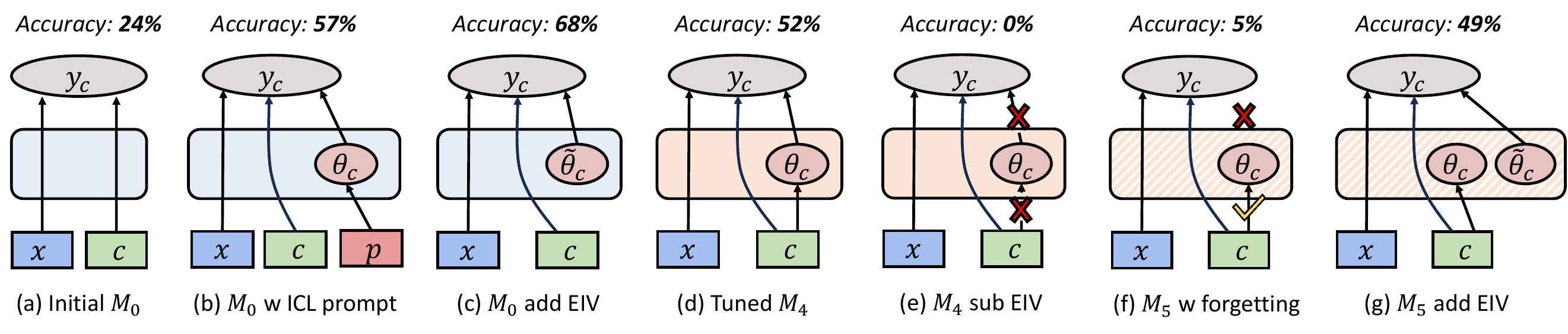}
  \vspace{-1.1em}
  \caption{Illustration of the instruction vector hypothesis. Here, $x$ represents the context, $c$ stands for a specific instruction, $y_c$ is the desirable output, and $\theta_c$ denotes the instruction vector. From (a) to (g), it visually details how these variables interact under different model conditions, with the accuracy above correlating to the respective performance on the CommonsenseQA task. The model configuration depicted in (d) is identified as the best state.}
  \label{fig:sec3:casual}
  \vspace{-1.1em}
\end{figure*}

Task in instruction dataset $D^{c}$ is to predict a target variable $y_c$, given a token sequence $x$ conditioned on instruction $c$. We assume a potentially high-dimensional latent variable $\theta_c$ exists, which governs the model's capability in following instruction $c$. This suggests a direct computational graph relationship among $x$, $c$, $\theta_c$, and $y_c$, mathematically depicted as $f_M(x, c, \theta_c) \rightarrow y_c $, as illustrated in Fig.~\ref{fig:sec3:casual}. Here, $f_M$ denotes the mapping function with model $M$ and we call $f_M(x, c, \theta_c) \rightarrow y_c $ the IV-associate computation graph.

Our hypothesis about the computational graph is supported by key observations illustrated in Fig.~\ref{fig:sec3:casual}: i) In (a-c), by intervening zero-shot input inference with representations drawn from in-context learning (ICL) samples (see Sec.~\ref{sec3.2}), accuracy improve from 24\% to 68\%. The effectiveness of this representation aligns with our definition of $\theta_c$, which may be activated by introducing a prompt before input or directly adding to the hidden states during the inference. ii) In (d,e), removing certain representations from well-behaved model results in a dramatic decline in performance from 52\% to 0\%, indicating a reliance on $\theta_c$ for producing $y_c$, beyond just the inputs $x$ and $c$. iii) Moreover, the differential impact on task performance in knowledge and instruction form point to a separation in the model's ability to handle $x$ and $c$. Hence, it's reasonable to conjecture that output relies on $\theta_c$ as opposed to $\theta_{x,c}$. Given the focus of this paper on instruction forgetting, the potential influence of $\theta_x$ is omitted in the following analysis.

\subsection{Instruction Vector}
\label{sec3.2}
We next consider how to extract $\theta_c$ for a given dataset $D^{c}$, drawing on the concept of function vectors proposed by~\citet{todd2023function}. This extraction is carried out using in-context learning (ICL) samples, where the model incorporates task-relevant information into its hidden states as it engages with examples with the ICL prompt. This process is associated with the emergence of $\theta_c$~\citep{todd2023function, hendel2023context}. Subsequently, a causal mediation analysis~\citep{Pearl2013InterpretationAI, NEURIPS2020_92650b2e, li2024understanding} is conducted on the ICL inputs to identify attention heads with significant causal impacts on the output, and aggregating their representations results in $\theta_c$. Interestingly, this vector remains effective even under zero-shot input scenarios, as demonstrated in Fig.~\ref{fig:sec3:casual} b,c. The detailed procedure is outlined below:


First, we start by gathering the task-conditioned activation for each model head by averaging the ICL input representation of the given task $D^{c}$, i.e., 
\begin{equation}
    \bar{h}_{l j}^c=\frac{1}{\left|D^{c}\right|} \sum_{(x_i, c) \in D^{c}} h_{\ell j}\left([p_i, x_i, c]\right).
\end{equation}
Where $p_i = [(x_1, c, y^c_1), ..., (x_N, c, y^c_N)]$ represents the N-shot ICL prompt text made up of held-out samples of task $c$, ${h}_{lj}$ is the model activation at the last token, layer $l$ and position $j$, and $\bar{h}_{lj}^c$ represents the task-conditioned activations.

Then to assess the existence of a cause-and-effect relationship between $\bar{h}_{l j}^c$ and correct output, we employ causal mediation analysis. The model will run on a counterfactual ICL input $[\hat{p_i},x_i,c]$ incorporating a label-shuffled prompt $\hat{p_i}=[(x_1, c, \hat{y}^c_1), ..., (x_N, c, \hat{y}^c_N)]$, typically leading to incorrect outcomes.
We then substitute the value of the specific head with the task-specific conditioned activation $\bar{h}_{lj}^c$ and calculate its causal effect (CE) on the model's output.
\begin{equation}
\begin{aligned}
\operatorname{CE}_{lj}([\hat{p_i},x_i,c])=P(y_{i}^c \mid [\hat{p_i},x_i,c] , M_{h^c_{lj}\rightarrow \bar{h}_{lj}^c}) \\ -P(y_{i}^c \mid [\hat{p_i},x_i,c] , M).
\end{aligned}
\end{equation}
Here, $M_{h^c_{lj}\rightarrow \bar{h}_{lj}^c}$ denotes the model with a replacement operation on attention head $(l,j)$ at last token of the input sentence. A higher CE suggests that the specific head's state is crucial in enabling accurate predictions, denoting the encoding of more task-relevant information.
For each head at layer $l$ and position $j$,we adopt the approach proposed by ~\citet{todd2023function} to calculate the average CE across a variety of tasks. Subsequently, we identify the top 10 heads with the highest average CE (recorded as set $\mathcal{S}$) as the most critical in conveying task-relevant information. The task vector $\theta_c$ is is then obtained by aggregating the task-conditioned activation from the attention heads in the set $\mathcal{S}$, i.e., $\theta_c=\sum_{a_{\ell j} \in \mathcal{S}} \bar{h}_{lj}^c$. 


We then evaluates the effectiveness of the Instruction Vector ($\theta_c$) through intervention experiments on the initial model across multiple datasets. The detail experiments can be found in Appendix~\ref{app:iv}. Results show that the IV significantly influences the output behavior for specific tasks, with its introduction notably improving zero-shot performance in certain tasks and removal diminishing the model's ability to produce correct outputs. This suggests that the model's specific abilities can be identified and analyzed by studying the corresponding IV.

In conclusion, our analysis suggests that forgetting in large language models (LLMs) results from a dynamic conflict between the dominance and suppression of existing computation graphs and new, specialized reasoning patterns learned from fine-tuning. This extends previous findings~\citet{kotha2024understanding} by utilizing IV framework to explore the underlying processes of forgetting in these models and confirming its theoretical underpinnings.




\section{Refinement of Training Methods to Mitigate Forgetting in LLMs}

Our previous research highlighted the critical role of the Instruction Vector (IV)-associated computation graph in Large Language Models (LLMs), crucial for maintaining the model's original capabilities. 
This insight prompted a reassessment of the training approaches to minimize forgetting. 
In this section, we show that fine-tuning guided by the \textit{Instruction Vector} helps balance the model's existing capabilities with new learning. 
This led us to reevaluate our training methods to prevent forgetting. 
This method, combined with existing continual learning algorithms, effectively reduces the forgetting of general abilities while preserving in-context reasoning capabilities, with minimal impact on plasticity.



\begin{table*}[]
\begin{center}
\begin{scriptsize}
\begin{tabular}{l|lll|lll|lll}
\toprule
\multirow{2}{*}{\textbf{Method}} & \multicolumn{3}{c|}{\textbf{TRACE}} & \multicolumn{3}{c|}{\textbf{LONG}} & \multicolumn{3}{c}{\textbf{FUNC}} \\ 
 & $HP$ & $IP$ & $OP$ &  $HP$ & $IP$ & $OP$ &  $HP$ & $IP$ & $OP$ \\\midrule\midrule
 \multicolumn{1}{l|}{Init} & 52.76 & 54.31 & 18.68 & 52.76 & 54.31 & 42.62 & 52.76 & 54.31 &  11.70 \\ \midrule

IncLora & 48.69 & 26.73 & 47.60 & 50.28 & 49.75 & 78.11 & 53.12 & 51.78 & 43.34 \\ 
 \multicolumn{1}{r|}{\textbf{+ IVG}} & 54.75 \textbf{(+6.06)} & 45.85 \textbf{(+19.1)} & 47.20 & 52.54 \textbf{(+2.26)} &	51.64  \textbf{(+1.89)} & 77.41 & 54.36	\textbf{(+1.24)} &	53.89	\textbf{(+2.11)} &	69.48 \\ \midrule
Ewc & 52.80 & 43.96 & 47.70 & 45.83 & 43.61 & 73.62  &  52.05	&	50.33	&	38.46  \\
 \multicolumn{1}{r|}{\textbf{+ IVG}} & 54.94 \textbf{(+2.14)}&54.58 \textbf{(+10.6)}&46.69& 52.38 \textbf{(+6.55)}& 53.36 \textbf{(+9.75)}	& 71.71  & 54.22	\textbf{(+2.17)} &	54.03	\textbf{(+3.70)} &	38.56\\ \midrule
Lwf & 52.71 & 54.44 & 34.68 &  53.29 &	54.29  & 69.39 &  53.33 & 54.43	&	57.91\\
 \multicolumn{1}{r|}{\textbf{+ IVG}} & 52.93 \textbf{(+0.22)}& 54.49 \textbf{(+0.05)} & 34.65 & 53.85 \textbf{(+0.56)} &	53.89 \textbf{(-0.40)}& 70.60 & 
53.59 \textbf{(+0.26)} & 54.23 \textbf{(-0.20) }& 61.92 \\ \midrule
OLora & 36.68 & 26.48 & 38.22 & 50.07 & 45.87 & 77.68  & 
54.13	& 52.38 & 42.12  \\
 \multicolumn{1}{r|}{\textbf{+ IVG}} &  49.08 \textbf{(+12.4)} & 46.35 \textbf{(+19.9)} & 39.78 & 52.05 \textbf{(+1.98)} & 51.48 \textbf{(+5.61) }&  76.98&  
53.94 \textbf{(-0.19)} &	53.90 \textbf{ (+1.52) }&58.13 \\ 
		
 				
 \bottomrule
\end{tabular}
\caption{Performance of baseline and their improved version with Instruction Vector Guided (\textbf{IVG}) training on three benchmarks (all results reported in this paper are averaged over 4 random seeds).}
\vspace{-0.7em}
\label{table:main}
\end{scriptsize}
\end{center}
\end{table*}

\paragraph{Instruction vector guided fine-tuning.}
In our analysis, we established a direct link between the IV-associated computation graph and the model's inherent task processing abilities. Forgetting typically occurs when the model's output becomes independent of the computation graph post-tuning. To address this, we propose an IV-guided training mechanism aimed at preserving capabilities before and after fine-tuning:

Initially, to utilize of the capabilities introduced by the IV, we propose a progressive IV intervention training. At training's start, the IV is explicitly included, with its influence gradually diminishing from 1 to 0 as training advances. This inclusion helps the model adhere to the computation graph outlined earlier, thus mitigating the overshadowing of existing capabilities by new learning. The original training objective is reformulated as:
\begin{equation}
\min_{M} \frac{1}{N} \sum_{i=1}^{N} \ell \left(y_i, M_{h^c_{lj}\rightarrow h^c_{lj}+s*\bar{h}^c_{lj}}(c, x_i)\right),
\end{equation}
where $M_{h^c_{lj}\rightarrow h^c_{lj}+s*\bar{h}^c_{lj}}$ denotes the intervention model on the causal attention heads set i.e., $(l,j) \in \mathcal{S}$. $s$ is a scaling factor that gradually decreases from 1 to 0 during training.

Furthermore, we introduce an IV-based KL-divergence loss function to better align the behaviour of fine-tuned computation graph with the IV indications:
\begin{equation}
\begin{aligned}
    \ell_{KL} = -KL[&P(y^c|[c,x],  M) \Vert \\ 
    & P(y^c| [c,x], M_{h^c_{lj}\rightarrow h^c_{lj}+\bar{h}^c_{lj}} )].
\end{aligned}
\end{equation}
This IV-guided fine-tuning approach leverages the existing knowledge within the model to direct the fine-tuning process, ensuring that the model retains a robust computation graph after fine-tuning and minimizes the impact of newly introduced knowledge on past knowledge and abilities.

\paragraph{Experimental Setup.}
Following the continual instruction tuning setup in Sec.~\ref{sec2}, we test our newly proposed method on TRACE and FUNC benchmarks additionally with a LONG sequence continual learning benchmark~\cite{razdaibiedina2023progressive} with 15 tasks. For the held-out evaluation set, we utilize Hellaswag, ARC-challenge, CommonsenseQA, and MMLU-social.
The experiments were conducted on the Llama2-7B-chat model, demonstrating its effectiveness in combination with existing continual learning methods, such as incremental Lora~\citep{hu2021lora} (\textbf{IncLora}), Learning without forgetting~\citep{li2017learning} (\textbf{Lwf}), Elastic weight consolidation~\citep{Kirkpatrick_2017} (\textbf{Ewc}), Orthogonal Lora~\citep{wang2023orthogonal} (\textbf{OLora}).
In our comparison, we prioritized training with hyper-parameters mentioned in previous works. We loaded the base LM into torch.bfloat16 to save memory and ran the experiments on 4 NVIDIA A100 GPUs.

To evaluate the performance of proposed algorithms, we utilize the average zero-shot held-out performance $HP = \frac{1}{n}\sum_{i=1}^{n} a^{h_i}_{T}$ to measure shift in general capabilities, average in-content held-out performance $IP =\frac{1}{n}\sum_{i=1}^{n} \hat{a}^{h_i}_{T}$ to evaluate forgetting in reasoning abilities, and overall training performance $OP = \frac{1}{T}\sum_{i=1}^{T} a^{t_i}_{T}$ to assess the degree of catastrophic forgetting on newly learned abilities. Here, $a^{i}_{j}$ represents the zero-shot evaluation score on the evaluation task $i$ after sequentially learning the $j$-th task. $\hat{a}$ denotes the in-context evaluation score. $h_i$ and $t_i$ denotes the $i$-th held-out evaluation set and $i$-th training task, respectively.

\paragraph{Results.} Table~\ref{table:main} shows the continual instruction tuning performance on three benchmarks, leading to several key observations:


\textit{Observation 1}: IV-guided training significantly prevents the loss of general and reasoning capabilities. Unlike most continual learning methods, which struggle with substantial forgetting of general abilities, our IV-guided training effectively mitigates this issue, resulting in an average forgetting rate on $HP$ of -0.16, compared to 5.03. Additionally, it enhances in-context performance from 37.90 to 50.05, underscoring the benefits of maintaining the computation graph.


\textit{Observation 2}: IV-guided training does not compromise the plasticity in learning new tasks. This approach shows only a slight reduction in the $OP$ metric, with changes of -0.03 and -0.55 for TRACE and LONG, respectively. This is in sharp contrast to the Lwf algorithm, which significantly reduces adaptability, resulting in a dramatic 12.92 drop in $OP$ on TRACE compared to IncLora.


\textit{Observation 3}: 
The likelihood of forgetting general abilities increases with the complexity of learning tasks. The benchmarks in Table~\ref{table:main}, ranked from simplest to most complex—FUNC, LONG, TRACE—show escalating HP forgetting rates from -0.40 to 2.89 and then to 5.04. The IV-guided training method effectively manages tasks across varying complexities, demonstrating its robustness in handling different learning challenges.




\section{Related work}

\paragraph{Catastrophic forgetting in fine-tuned language models.}
Fine-tuning foundational LLMs~\citep{touvron2023llama_1,touvron2023llama} has become a generic technique for enhancing their capacity of following instructions~\citep{wei2022finetuned,zhang2024llamaadapter,zhang2024instruction} and mastering domain-specific content~\citep{yue2023disclawllm,christophe2024med42}. 
However, adopting such technique can have a negative effect of hurting the original ability of LLMs, which is widely known as Catastrophic Forgetting~\citep{Kirkpatrick_2017,zhai2023investigating,luo2024empirical,kotha2024understanding,wu2024continual}.
In context of LLMs, existing approaches towards mitigating this issue can mostly be categorized into three types: regularizing the update of model parameters~\citep{Kirkpatrick_2017,huang-etal-2021-continual,cha2021cpr}, replaying previous or self-synthesized data~\citep{scialom-etal-2022-fine,huang2024mitigating} and resisting interference via parameter-efficient fine-tuning~\citep{razdaibiedina2023progressive,wang2023orthogonal}.
\paragraph{Mechanistic analysis to fine-tuning.}
Existing works on analyzing the internal mechanism~\citep{räuker2023transparent,ferrando2024primer} of fine-tuning mainly focus on the question that how LLMs acquire new capacity in the learning process, arguing that models learn a minimal transformation on top of the original capability~\citep{jain2024mechanistically} (wrappers), subtractable and reusable parameter shift vectors~\citep{huang2024chat,gao2024ethos} (task vectors) and to align input queries with their internal knowledge that are already acquired in the pre-training stage~\citep{ren2024learning}. 
Nevertheless the inherent reason for the forgetting issue brought by fine-tuning currently remains unclear, and hence our work instead targets on this important point.

\section{Conclusion}

In our study, we introduce Instruction Vector (IV), which enables detailed analysis of LLMs task processing capabilities. By analyzing IV dynamics before and after training, we show that forgetting is caused by the overlay of new reasoning patterns over pre-existing skills, while the performance can be recovered by adding the IV to the computation graph. Additionally, our proposal of IV-guided training as a fine-tuning method successfully reduces forgetting by maintaining harmony between the model's computation graph and the IV-associated one. These findings offer valuable insights into the internal mechanisms causing forgetting in LLMs and are expected to contribute to advancing the development and application of LLMs alignment.


\section{Limitation}

The IV-guided training method does not directly address the problem of forgetting newly learned knowledge in most cases, and needs to be combined with existing continual learning methods to acquire this ability. This is because we overcome forgetting by preserving the computation graph, which indicates the existing capabilities, making it unable to protect newly acquired knowledge. Interestingly, in the FUNC dataset, our method significantly reduced forgetting of new knowledge on IncLora and OLora. These tasks have simple and deterministic instructions, which may allow the model to integrate new capabilities with the constructed computation graph during IV-guided training, thus overcoming forgetting. This inspires us to investigate the adaptability and generalization of the computation graph in future research for more refined learning of new knowledge.

Second, we aggregate attention heads to extract the Instruction vector in this paper. Although this method is fast and efficient, it is susceptible to input noise and may suffer from insufficient expressiveness. Therefore, we plan to use optimization-based methods in future to extract a more generalized and accurate Instruction vector.

Finally, due to limitations in experimental resources, we did not conduct experiments on multiple backbones. In the future, we will validate our hypothesis about forgetting on more LLMs.

\bibliography{custom}

\newpage
\appendix

\section{Datasets}
\label{app:dataset}

Three continual instruction tuning benchmarks and severel general evaluation datasets are adopts in this paper. The detailed information is as follows:

\paragraph{TRACE benchmark.}

TRACE benchmark is released by~\citet{wang2023trace} for the study of forgetting in LLMs, which consists of 8 different complex generation tasks including multi-choice QA, code generation, mathematical reasoning and summary. Without loss of generaliztion, we select 6 out of 8 raw tasks to construct the training sequence as our experiments setup. The statistical information is listed in Table~\ref{table:trace}, while order in Table~\ref{table:order}

The training epoch for this benchmark is 5 for C-STANCE, Py150, NumGLUE-cm, 3 for FOMC and ScienceQA, and 7 for MeetingBank. We evaluate them with a self-construct evaluation code based on OpenCompass code framework.

\begin{table*}[]
\vspace{-0.8em}
\begin{center}
\begin{small}
\begin{tabular}{l|llllll}
\toprule Dataset & Source & Category & Avg len & Metric & Language & \#data \\ \midrule \midrule
ScienceQA & Science & Multi-Choice QA & 210 & Accuracy & English & 5,000 \\  
FOMC & Finance & Multi-Choice QA & 51 & Accuracy & English & 5,000 \\ 
MeetingBank & Meeting& Summary & 2853 & ROUGE-L & English & 5,000 \\ 
C-STANCE & Social media& Multi-Choice QA & 127 & Accuracy & Chinese & 5,000 \\ 
Py150 & Github& Code generation & 422 & Edim similarity & Python & 5,000 \\ 
NumGLUE-cm & Math & Math reasoning & 32 & Accuracy & English & 5,000 \\
\bottomrule
\end{tabular}
\caption{A summary of dataset statistics in TRACE includes information on the source of the context, average length in terms of word count for English, German, and code datasets, and character count for Chinese.}
\label{table:trace}
\end{small}
\end{center}
\end{table*}

\paragraph{LONG benchmark.}
LONG benchmark is widely utilized in existing continual learning works~\citet{wang2023orthogonal, razdaibiedina2023progressive} with 15 task. The training epoch is set to 1 for each task following ~\cite{wang2023orthogonal}. The statistical information is listed in Table~\ref{table:long}.

\begin{table*}[]
\vspace{-0.8em}
\begin{center}
\begin{small}
\begin{tabular}{l|llllll}
\toprule Dataset & Source & Category & Avg len & Metric & Language & \#data \\ \midrule \midrule
Yelp & Yelp reviews & Sentiment analysis & 757  & Accuracy  & English& 5,000 \\
SST2 & Movie reviews & Sentiment analysis & 62 & Accuracy& English & 2,000 \\
Amazon & Amazon reviews & Sentiment analysis &  458 & Accuracy & English& 5,000 \\
IMDB & Movie reviews &  Sentiment analysis & 1,340 & Accuracy & English& 2,000\\
DBpedia & Wikipedia & Topic classification & 324 & Accuracy & English & 14,000 \\
Yahoo & Yahoo Q\&A  & Topic classification & 562 & Accuracy & English& 10,000 \\
AG News & News & Topic classification & 259  & Accuracy & English& 4,000\\
WiC & Lexical database & Disambiguation & 93 & Accuracy& English &2,000 \\
QQP & Quora & Paraphrase &  158 & Accuracy & English& 2,000 \\
RTE & News, Wikipedia & NLI & 365 & Accuracy & English& 2,000 \\
MNLI & Multi & NLI & 205 & Accuracy & English& 3,000 \\
CB & Multi & NLI & 365 & Accuracy & English& 250\\
COPA & blogs, encyclopedia & Question answering & 161 & Accuracy & English & 400  \\
BoolQ & Wikipedia & Question answering & 655  & Accuracy  & English& 2,000 \\
MultiRC & SuperGLUE & Question answering & 1728 & Accuracy & English& 2,000\\

\bottomrule
\end{tabular}
\caption{A summary of dataset statistics in LONG.}
\label{table:func}
\end{small}
\end{center}
\end{table*}

\paragraph{FUNC benchmark.} FUNC benchmark is adapted from the datasets in ~\citet{todd2023function}, in which tasks have clear and simple instructions. For example, task Verb-Spanish and Last-Spanish are both translation task but differ in the selection from list. The training epoch is set to 10 for each task. The statistical information is listed in Table~\ref{table:long}.

\begin{table*}[]
\vspace{-0.8em}
\begin{center}
\begin{small}
\begin{tabular}{l|llllll}
\toprule Dataset & Source & Category & Avg len & Metric & Language & \#data \\ \midrule \midrule
Alphabetically\_last\_of\_5 & $- $& Extractive, Capital & 144 & Accuracy  & English& 700 \\
Choose\_last\_of\_5\_spanish & $-$ & Extractive, Translation & 109 & Accuracy& English, Spanish & 700 \\
AG News & News & Topic classification,QA &  285 & Accuracy & English& 1,500 \\
Object\_v\_concept\_5\_spanish & $-$&  Extractive, Translation & 106 & Accuracy& English, Spanish& 700\\
Verb\_v\_adjective\_5\_spanish& $-$ &  Extractive, Translation & 106 & Accuracy& English, Spanish& 700\\
Sentiment & $-$&  Sentiment analysis,QA & 75 & Accuracy& English &816\\
\bottomrule
\end{tabular}
\caption{A summary of dataset statistics in FUNC.}
\label{table:long}
\end{small}
\end{center}
\end{table*}

\paragraph{General evaluation sets.} For the general evaluation datasets, we utilize Hellaswag~\citep{zellers2019hellaswag}, ARC-challenge~\citep{clark2018think}, CommonsenseQA~\citep{talmor2018commonsenseqa}, and MMLU-social~\citep{hendrycks2020measuring}. The datasets is downloaded from \url{https://github.com/open-compass/opencompass} and evaluate with the OpenCompass code framework.

\section{Input template}
In this paper, the instruction template is divided into two parts, refer to Sec.~\ref{sec2}. The first part corresponds to knowledge probability, namely $(x, y)$, and the second part corresponds to Instruction probability, namely $(x, c, y^c)$. The specific template content used for each dataset is given below, as show in Table~\ref{table:template_long}, Table~\ref{table:template_know}, and Table~\ref{table:template_inst}.

\begin{table*}[]
\vspace{-0.8em}
\begin{center}
\begin{small}
\begin{tabular}{c|l}
\toprule 
 Task & \multicolumn{1}{c}{ Template } \\ \midrule \midrule
 Yelp, SST2, Amazon, IMDB & \begin{tabular}{l} "Input": "What is the sentiment of the following paragraph? [$x$] \\ Choose one from the option.", "Output": "[$y$]"\end{tabular}  \\\midrule
DBPedia, Yahoo, AG NEws &\begin{tabular}{l} "Input": "What is the topic of the following paragraph? [$x$] \\ Choose one from the option.", "Output": "[$y$]"\end{tabular} \\ \midrule
 QQP & \begin{tabular}{l} "Input": "Whether the [$x_1$] and the [$x_2$] have the same meaning? \\ Choose one from the option.", "Output": "[$y$]" \end{tabular} \\ \midrule
 RTE, MNLI, CB & \begin{tabular}{l} "Input": "What is the logical relationship between the [$x_1$] and the [$x_2$]? \\Choose one from the option.", "Output": "[$y$]"\end{tabular} \\ \midrule
BoolQA & \begin{tabular}{l} "Input": "According to the following passage, is the question true or false? [$x$] \\ Choose one from the option.", "Output": "[$y$]" \end{tabular} \\ \midrule
MultiRC & \begin{tabular}{l}  "Input": "According to the following passage and question, is the candidate answer true  \\ or false?  [$x$]  Choose one from the option.", "Output": "[$y$]"  \end{tabular} \\ \midrule
WiC & \begin{tabular}{l}  "Input":  "Given a word and two sentences, whether the word is used with the same sense \\ in both sentence? Choose one from the option.",  "Output": "[$y$]" \end{tabular} \\
\bottomrule
\end{tabular}
\caption{Input template for tasks in LONG benchmark.}
\label{table:template_long}
\end{small}
\end{center}
\end{table*}

\begin{table*}
    \centering 
    \begin{small}
    \begin{tabular}{c|l}
    \toprule
    Benchmark & Task Sequence \\
    \midrule \midrule
    TRACE & C-STANCE$\rightarrow$ FOMC$\rightarrow$ MeetingBank$\rightarrow$ Py150$\rightarrow$ ScienceQA $\rightarrow$ NumGLUE-cm \\ \midrule
    LONG & Yelp$\rightarrow$ Amazon$\rightarrow$ MNLI$\rightarrow$ CB$\rightarrow$ COPA$\rightarrow$ QQP$\rightarrow$ RTE$\rightarrow$ IMDB$\rightarrow$\\
    & SST2$\rightarrow$ DBpedia$\rightarrow$ AG News$\rightarrow$ Yahoo$\rightarrow$ MultiRC$\rightarrow$ BoolQ$\rightarrow$ WiC\\ \midrule
    FUNC &  Verb-Spanish $\rightarrow$ Last-Spanish$\rightarrow$ Sentiment-mc$\rightarrow$ Object-Spanish$\rightarrow$ Alphabetically-Capital $\rightarrow$ AGNews-mc \\
    \bottomrule
    \end{tabular}
    \end{small}
    \caption{The orders used for each benchmark.}
    \label{table:order}
    
\end{table*}

\begin{table*}[]
\vspace{-0.8em}
\begin{center}
\begin{small}
\begin{tabular}{c|l}
\toprule 
 Task & \multicolumn{1}{c}{ Prompts } \\ \midrule \midrule
ScienceQA & \begin{tabular}{l} "Input": [$x$] "Output": [$y$]\end{tabular} \\ \midrule
FOMC & \begin{tabular}{l} "Input": “Text: [$x$] The monetary policy stance of above text is “, "Output": “[$y$]” \end{tabular} \\ \midrule
C-STANCE & \begin{tabular}{l} (Translate Chinese to English) "Input": ”Text: [$x_1$] Object: [$x_2$] \\ The attitude of above text towards object is”, "Output": “[$y^c$]"\end{tabular} \\ \midrule\midrule
Last-Spanish & \begin{tabular}{l}
"Input": "Choose the last item in the list. [$x$]", "Output": "[$y^c$]"\end{tabular} \\ \midrule
Object-Spanish & \begin{tabular}{l} "Input": "Choose the object in the list. [$x$]",
”Output": "[$y$]" \end{tabular} \\ \midrule
Verb-Spanish & \begin{tabular}{l}  "Input": "Choose the verb in the list. [$x$]",
"Output": "[$y$]"\end{tabular} \\ \midrule
Alphabetically-Capital & \begin{tabular}{l}
"Input": "Choose the last item in the order of alphabetically in the list. [$x$]",
"Output": "[$y$]"
\end{tabular} \\ \midrule
AGNews-mc & \begin{tabular}{l} "Input": "Classify the following news with the label Business, \\ Science, Sports, and World.  [$x$] ",
"Output": "[$y$]" \end{tabular} \\ \midrule
Sentiment-mc & \begin{tabular}{l} "Input": "[$x$]",
"Output": "[$y$]"\end{tabular} \\ 
\bottomrule
\end{tabular}
\caption{Input template for calculating knowledge probability for different tasks.}
\label{table:template_know}
\end{small}
\end{center}
\end{table*}

\begin{table*}[]
\vspace{-0.8em}
\begin{center}
\begin{small}
\begin{tabular}{c|l}
\toprule 
Task & \multicolumn{1}{c}{ Prompts } \\ \midrule \midrule
ScienceQA & \begin{tabular}{l} "Input": "Choose an answer for the following question and \\ give your reasons. Question: [$x$] Answer:", "Output": "[$y^c$]" \end{tabular} \\ \midrule
FOMC & \begin{tabular}{l} "Input": "What is the monetary policy stance for the following text? A. dovish, B. hawkish, \\ C. neutral. Choose one from A, B and C. Text: [$x$] Stance:", "Output": "[$y^c$]" \end{tabular} \\ \midrule
C-STANCE & \begin{tabular}{l} (Translate Chinese to English) "Input": ”Determine the attitude of \\ the following text towards the specified object. Select one: A. Support, \\ B. Oppose, C. Neutral. Output A, B or C. Text: [$x_1$] Object: [$x_2$] Attitude:”, "Output": “[$y^c$]"\end{tabular} \\ \midrule
MeetingBank & \begin{tabular}{l} "Input": "Write a summary of the following meeting transcripts. \\ Meeting transcripts: [$x$] Summary:", "Output": “[$y$]”\end{tabular} \\ \midrule
Py150 & \begin{tabular}{l} "Input":  “<s> [x]”, "Output": “[$y$]”\end{tabular} \\ \midrule
NumGLUE-cm & \begin{tabular}{l} "Input": "Solve the following math problem. Question: [$x$] Answer:”, "Output": “[$y$]”\end{tabular} \\ \midrule\midrule
Last-Spanish & \begin{tabular}{l} 
        "Input": "Choose the last item in the list and \\ translate to spanish. [$x$]",
    "Output": "[$y^c$]"\end{tabular} \\ \midrule
Object-Spanish & \begin{tabular}{l} "Input": "Choose the object in the list and \\ translate to spanish. [$x$]",
    "Output": "[$y^c$]" \end{tabular} \\ \midrule
Verb-Spanish & \begin{tabular}{l}  "Input": "Choose the verb in the list and  \\ translate to spanish. [$x$]",
    "Output": "[$y^c$]"\end{tabular} \\ \midrule
Alphabetically-Capital & \begin{tabular}{l} 
        "Input": "Choose the last item in the order of alphabetically in the list and \\ print in the capital form. [$x$]",
    "Output": "[$y^c$]"
     \end{tabular} \\ \midrule
AGNews-mc & \begin{tabular}{l} "Input": "[$x$] A: Business B: Science C: Sports D: World",
    "Output": "[$y^c$]" \end{tabular} \\ \midrule
Sentiment-mc & \begin{tabular}{l} "Input": "[$x$] a: positive b: negative",
    "Output": "[$y^c$]"\end{tabular} \\ 
\bottomrule
\end{tabular}
\caption{Input template for calculating instruction probability and training for different tasks.}
\label{table:template_inst}
\end{small}
\end{center}
\end{table*}

\section{Implementation}
\label{app:implement}
We adopt LLAMA2-7B-Chat~\citep{touvron2023llama} as the base model, with its effectiveness in both understanding world knowledge and following instructions. Without specific notification, the model is fine-tuned with LORA approach~\cite{hu2021lora}, where the rank dimension set to 8 and the target module is query and value weight matrices. For IncLora, OLora, and Lwf methods, a new adapter is initialized at the beginning of learning new task while keep the previous Lora adapters fixed. For Ewc, only one big adapter is initialized during the sequential learning, where rank is set to 48 for TRACE and FUNC, and 60 for LONG.

The maximum input sequence length is set to 512 and the maximum output sequence length is set to 50. We train the model with the decoder only task calculating gradient only on the output tokens. We use an Adam optimizer with a weight decay of 0.01 and the learning rate set to 1e-4 for TRACE and FUNC, 1e-3 for LONG (following ~\cite{wang2023trace}). The batch size is set to 8 and accumulate gradient step is set to 2 for each GPU while we run on 4 A100 GPUs with Deepspeed. The training size and epochs can be found in the introduction of datasets. 

As for the hyperparameters, we perform a grid search on the scale of KL-divergence loss within [1, 0.5, 0.25, 0.05, 0.01] and set 0.05 as the final choice. For the hyperparameters of existing continual learning methods, I refer to the well-searched value reported in previous paper.

\section{Implementation Detail of Instruction Vector Framework}

When extracting the Instruction Vector from in-context samples, we use 10-shot input prompt randomly selected from held-out training dataset. The task-conditioned activations are average on samples filtered with correct 10-shot answer from the validation set with 200 samples. As for the set $\mathcal{S}$ of the casual attention heads, we follow the position in ~\citet{todd2023function} and validate its efficiency on our own datasets. Specifically, the set $\mathcal{S}$ is $[(14, 1), (11, 2), (9, 25), (12, 15), (12, 28), (13, 7),\\$$(11, 18), (12, 18), (16, 10), (14, 16)]$.


\section{Effectiveness of Instruction Vector}
\label{app:iv}

\begin{figure}[h]
  \centering
  \vspace{-1em}
  \includegraphics[width=0.8\linewidth]{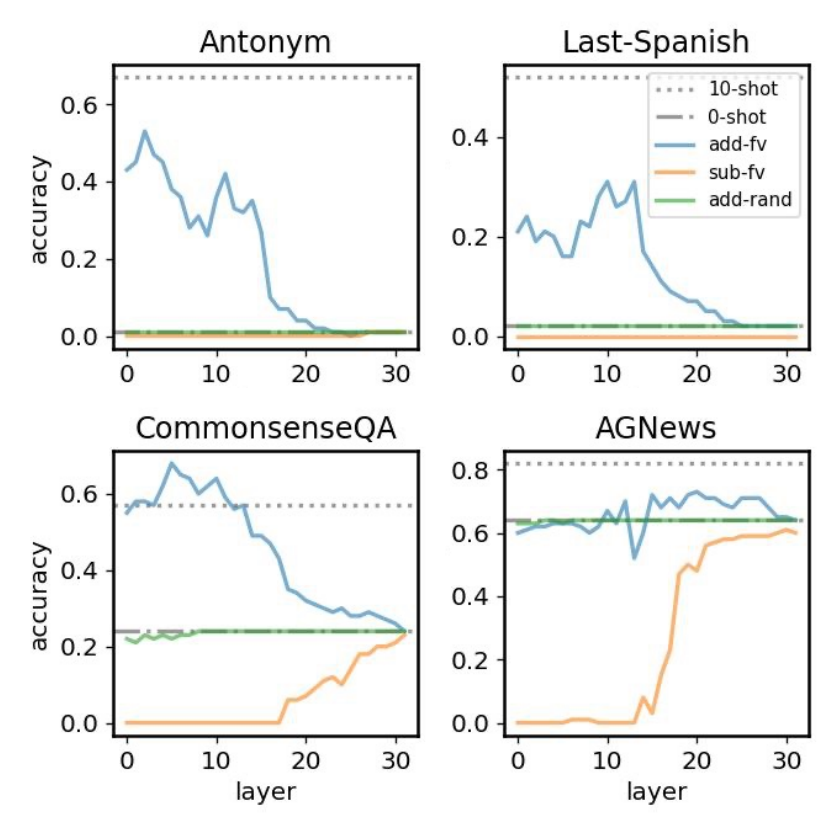}
  \vspace{-1em}
  \caption{Intervention results on four datasets via Enhanced Instruction Vector.}
  \label{fig:sec3:intervention}
\end{figure}

To assess the effectiveness of the extracted $\theta_c$, referred to as the Instruction Vector (IV) in this study, we conduct a series of intervention experiments across multiple datasets (see Fig.~\ref{fig:sec3:intervention}) on the initial model. These experiments consisted of either inserting or removing an IV at the hidden states of a specific layer at the the last token position, to examine the influence on the model output. More precisely, in the transformer's forward residual stream, the instruction vector $\theta_c$ modifies the hidden states at a select layer $l$ as $h_l = h_l + \theta_c$.

We reported the intervention findings on four distinct datasets: 1) CommensenseQA, multiple-choice questions on common sense reasoning; 2) Antonym, a task aimed at generating antonyms; 3) AGNews, a text classification task with the article's category as the label; and 4) Last-Spanish, a task that output the Spanish translation of the list's final item. 
The results highlighted that the IV directly affects the model's output behavior for specific tasks. In tasks such as Antonym, Last-Spanish, and CommonsenseQA, introducing IV significantly improved the zero-shot performance from a low level. Conversely, in the cases of AGNews and CommonsenseQA, removing the IV resulted in a deterioration of the model's ability to produce the correct output. In contrast, interventions with random vectors had a negligible effect on the model. These findings indicate that the specific capabilities of the model can be identified and analyzed by examining the dynamics of the corresponding IV.

\section{Recovery with Instruction Vector}
\begin{figure}[h]
  \centering
  \vspace{-1em}
  \includegraphics[width=0.6\linewidth]{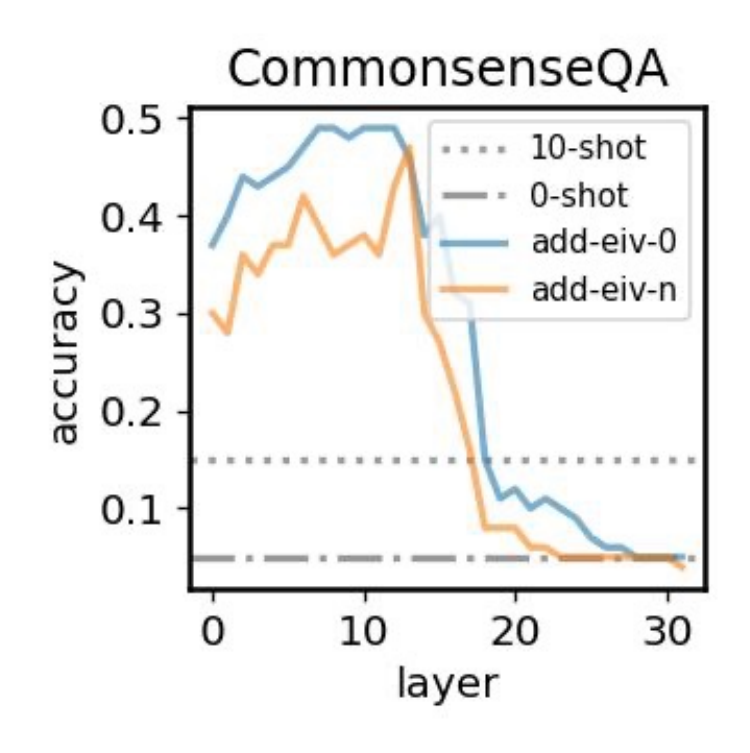}
  \vspace{-1em}
  \caption{The intervention results on model sequentially fine-tuned on TRACE benchmark.}
  \label{fig:app:intervention_aftertune}
\end{figure}

We conducted an intervention experiment on the CommonsenseQA task with the fine-tuned model on the TRACE benchmark (refer to Fig.~\ref{fig:app:intervention_aftertune}). The results show that the model exhibited significant forgetting in both 0-shot and 10-shot performance, dropping to 0.03 and 0.15, respectively. However, integrating IV into the model (as shown in Fig.~\ref{fig:sec3:casual}(g)), i.e., $h_l=h_l+\theta_c$, resulted in a substantial recovery in model performance. Performance reached 0.47 when using IV derived from the current model and 0.49 with IV from the initial model.

\end{document}